\documentclass[runningheads]{llncs}
\usepackage[T1]{fontenc}
\usepackage{graphicx}
\usepackage{amsmath}
\usepackage{url,hyperref,cleveref}
\usepackage{booktabs}
\usepackage{svg}
\usepackage[square,numbers,sort]{natbib}  

\begin{document}
\title{On when is Reservoir Computing with Cellular Automata Beneficial?}
%
%

\author{Tom Eivind Glover\inst{1}\orcidID{0000-0002-3489-8731} \and
Evgeny Osipov\inst{3}\orcidID{0000-0003-0069-640X} \and
Stefano Nichele\inst{1,2}\orcidID{0000-0003-4696-9872}}
\authorrunning{T. Glover et al.}
%

\institute{Department of Computer Science, Oslo Metropolitan University, Oslo, Norway 
\email{tomglove@oslomet.no}\\
\and
Department of Computer Science and Communication, Østfold University College, Halden, Norway
\email{stefano.nichele@hiof.no}\\
\and
Department of Computer Science, Electrical and Space Engineering, Luleå University of Technology, Luleå, Sweeden\\
\email{Evgeny.osipov@ltu.se}}
\maketitle              
\begin{abstract}
Reservoir Computing with Cellular Automata (ReCA) is a relatively novel and promising approach. It consists of 3 steps: an encoding scheme to inject the problem into the CA, the CA iterations step itself and a simple classifying step, typically a linear classifier. This paper demonstrates that the ReCA concept is effective even in arguably the simplest implementation of a ReCA system. However, we also report a failed attempt on the UCR Time Series Classification Archive where ReCA seems to work, but only because of the encoding scheme itself, not in any part due to the CA. This highlights the need for ablation testing, i.e., comparing internally with sub-parts of one model, but also raises an open question on what kind of tasks ReCA is best suited for.

\keywords{Cellular Automata  \and Reservoir Computing \and ReCA \and MNIST \and UCRArchive.}
\end{abstract}

\section{Introduction}
Reservoir Computing (RC) is a method that relies on an untrained reservoir and a linearly trained simple classifier, making it a very energy-efficient method to train. CA is a regular vector of discrete values that interact with boolean operations. Therefore, CA is relatively simple 
to implement in hardware, e.g., FPGA, making it energy-efficient in execution. When RC and CA are combined into Reservoir Computing with Cellular Automata (ReCA), it becomes, in theory, an energy-efficient method to both train and run. Such features are important for EdgeAI \citep{singh2023edge}. The property of energy efficiency can enable capabilities closer to or on the edge devices. This, in turn, can improve the system's privacy, cost, capabilities, and availability. 

Most ReCA works explore ReCA on time-dependent benchmarks, such as chaotic time series prediction or the 5-bit memory benchmark \citep{glover2023investigating, kantic2024relicada}. These benchmarks can only be solved if time is taken into account and the ReCA solutions are designed such that the CA is the only place it can be contained. This makes ablation testing infeasible. One can still demonstrate value by comparing to other known methods, but as different encoding schemes can react differently to different substrate and again to different classifiers, it is less clear how much value the CA reservoir is actually providing. 
Furthermore, the ReCA method has in a few cases also been explored on benchmarks such as MNIST \citep{garcia2020low, moran2018reservoir, liang2021bloomca} that do not depend on time. Therefore, there is a need to clarify what role the CA serves. As far as we can tell, only \citep{liang2021bloomca} did an ablation test, where accuracy was 89.58\% without and 91.86\% with CA. While these results are informative, it is worth mentioning that our trivial solution achieves higher accuracy on the MNIST benchmark.

This work's main contribution is exploring ReCA on MNIST, using a simple design and demonstrating that it works for many ECA rules. The simple design is a form of Minimum Viable Product (MVP), a design principle where only the vital features are implemented. This provides value as a feasibility study and clarifies that the CA can be useful. The simple design also serves as an example of ReCA that can easily be understood and replicated. Additionally, we show that MNIST can be solved relatively easily even if all the grayscale information is removed. 
We also explored the method on the UCR dataset, a benchmark of several time series classification tasks. As RC works well with time series, it seemed a natural benchmark to test on. We observed that it seems to work with a similarity-preserving encoding scheme. However, we show that the encoding scheme itself does all the work. Making this a great example of the importance of ablation testing.

\section{Background}
\subsection{Cellular Automata}
Cellular Automata (CA) are a simple model consisting of a grid of cells in a limited set of $k$ discrete states. The grid is uniformly connected, typically in 1 or 2 dimensions. The cell state changes iteratively, depending on the state of the neighbouring cells. The combination of the neighbour states deterministically defines the next state via a lookup table, typically called the Transition Table (TT). CA was first used to study self-replication by John von Neumann in 1940 but published in 1966 \citep{neumann1966theory}. It can be considered an idealised system for parallel and decentralised computation \citep{mitchell2001life}.
\subsubsection{Elementary Cellular Automata (ECA)}

\begin{figure}[h]
\begin{center}
\includegraphics[width=0.8\columnwidth]{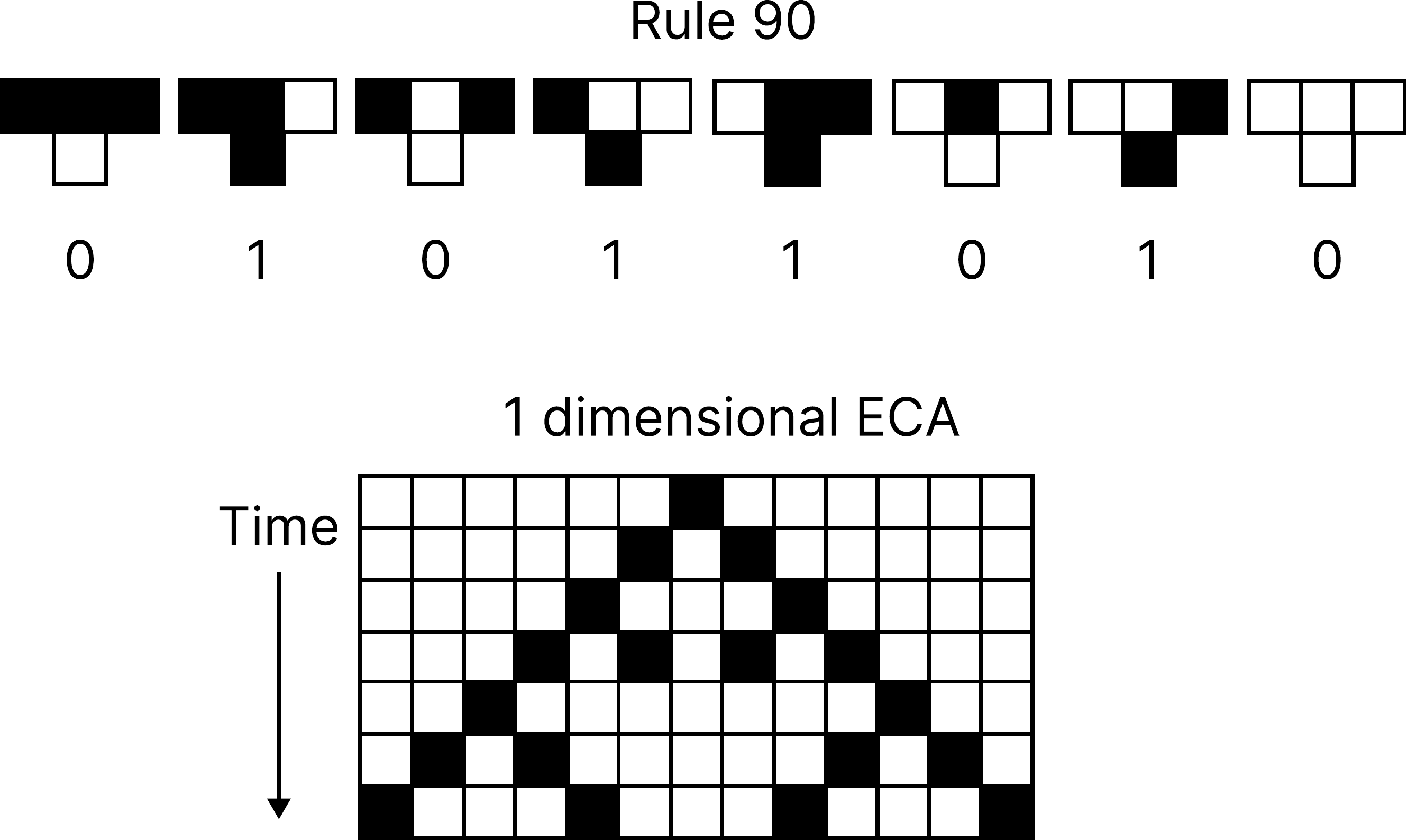} 
\caption{Example of ECA with rule 90 with TT, starting from a central cell on, executing 7 time-steps.}
\label{fig:ECAex}
\end{center}
\end{figure}

Elementary Cellular automata (ECA) is a subset of CA that has 1-dimension, binary states ($S=2$) and 3 neighbours ($K=3$) (left, right and centre). Therefore, ECA only has $S^{S^K} = 2^{2^3} = 256$ possible rules (unique TT), and the whole set of these rules is often named the rule space. It is a convention to name individual rules in a rule-space after the output states of the TT $\mbox{Binary}(01011010) = \mbox{Decimal}(90)$. CA is deterministic, and the rule, together with the initial condition, leads the CA into a set of subsequent states called the trajectory. An example of rule 90 can be seen in Figure \ref{fig:ECAex}. Rule 110 has even been shown to be computationally universal \citep{cook2004universality}, but one can question whether that is a useful definition of computation for a parallel and distributed computational substrate \citep{barbora2021hirarchy}. 

Due to symmetries in the rule-space, one can use trivial transformations to reduce the 256 ECA rules to just 88 that represent the entire rule-space, called the minimum equivalent rules. The trivial transformations, often called reflection and complement, were first pointed out in \citep[p.~51, p.~176]{walker_study_1965}, but was popularised and demonstrated for CA later in \citep{wolfram_theory_1986, wolfram2018tables, li_structure_1990, wuensche_global_1992}

\subsection{Reservoir Computing (RC)}
Reservoir Computing (RC) is a substrate-independent framework for computing. RC is independent because it works on many different substrates, but to be clear different substrates would of course have different capabilities. The RC framework consists of 3 parts, the encoding, the untrained reservoir and the output.

\begin{figure}[h]
\begin{center}
\includegraphics[width=0.5\textwidth]{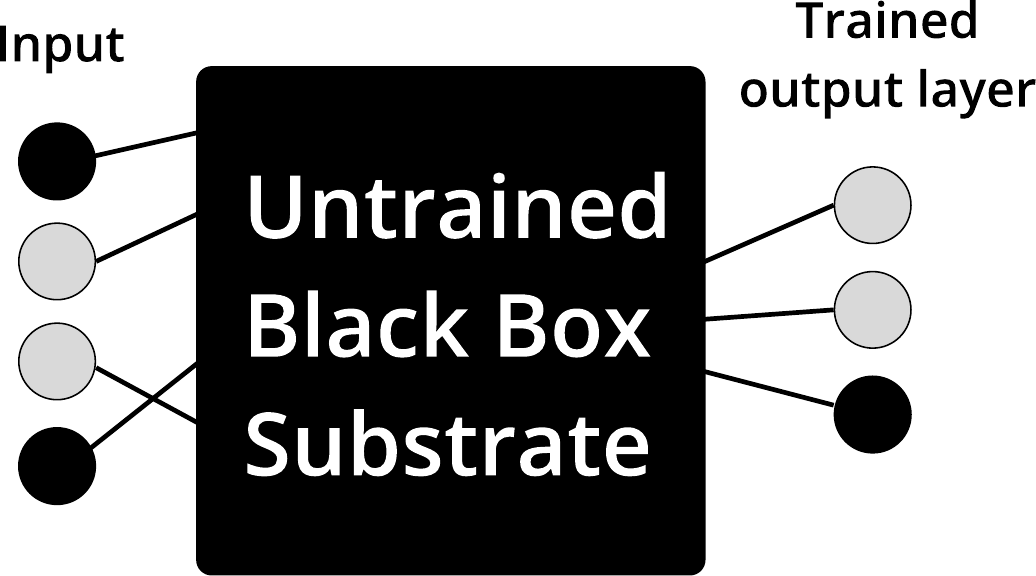}
\caption{RC as a substrate-independent framework }
\label{fig:RCBlackBoxModel}
\end{center}
\end{figure}

The encoding part inputs information into the untrained reservoir and typically into higher dimensions. The untrained reservoir typically expands, modifies or changes the information, but could, in the context of the framework, be considered a black box as seen in Figure \ref{fig:RCBlackBoxModel}. The output part is typically linear, does dimensional reduction and extracts useful features.

\begin{figure}[h]
\begin{center}
\includegraphics[width=0.5\textwidth]{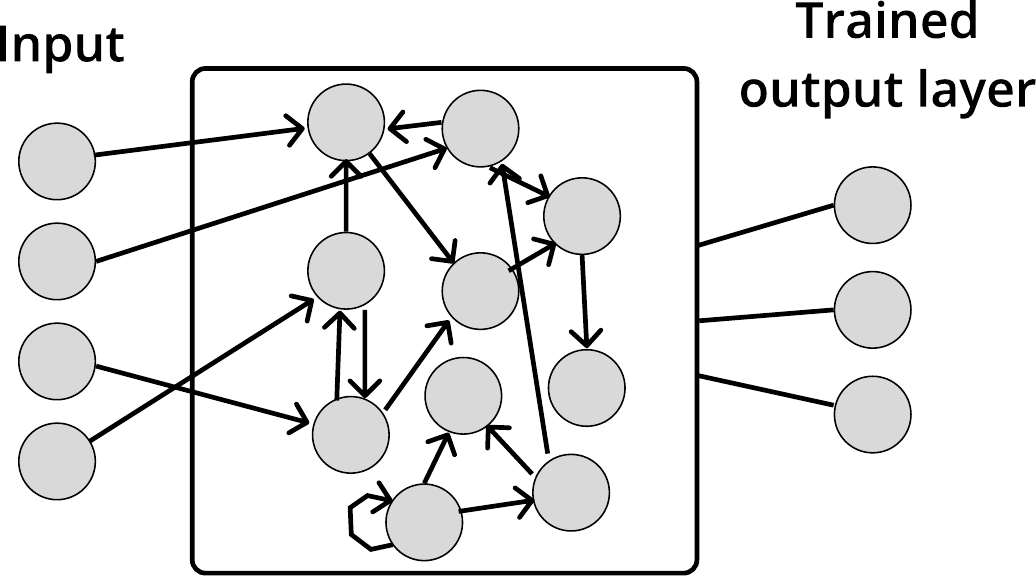}
\caption{Basic network Architecture of an ESN.}
\label{fig:ESNModel}
\end{center}
\end{figure}

The RC concept originated in echo state networks (ESN) using recurrent neural networks as a substrate \citep{jaeger2001echo} and in liquid state machines (LSM) using a spiking neural network for a substrate \citep{maass2002real}. Since then, both ESM and LSM and a host of other substrates have been categorised under the umbrella term of RC. Due to RC substrate-independent nature, many different substrates have been explored and compared  \citep{tanaka2019recent}. Some explore different topology configurations as in \citep{gallicchio2016deep}, where deep layered sub-reservoirs were analysed instead of the typical one large reservoir. RC is also a prevalent method with physical reservoirs \citep{tanaka2019recent}, as an extreme example in \citep{fernando2003pattern}, it was demonstrated that RC could use the surface waves on a bucket of water as a reservoir, and they successfully solved speech recognition and xor tasks using this substrate. One interesting substrate is real biological neural networks (BNN), specifically disassociated neurons that self-organise over a microelectronic array \citep{aaser2017towards}. 


\subsection{Reservoir Computing with CA (ReCA)}
The first study that introduced CA as a substrate in reservoir computing is \citep{yilmaz2014reservoir}. This study investigated Game of Life and several ECA rules as reservoir substrates and tested them on a 5-bit and 20-bit memory benchmark. In addition, it presents a theoretical comparison of CA vs ESN, using the metric of the number of operations needed to solve the benchmark, which documents a clear advantage of using CA.

As an ECA reservoir only relies on simple discrete binary interactions between cells, it affords a hardware-friendly implementation.
In \citep{moran2019energy}, ReCA using ECA with a max-pooling and softmax strategy was implemented on a Field Programmable Gate Array (FPGA).
In \citep{olin2019cellular}, a CA was implemented on Complementary metal-oxide-semiconductor (CMOS) combined with a custom hardware SVM implemented in resistive random-access memory (ReRAM).
In \citep{liang2021bloomca}, a synthesised hardware implementation of ReCA using ECA with a max-pooling and ensemble bloom filter classifier. It showed impressive results compared to "state-of-the-art" in terms of energy efficiency, memory usage and area(number of gates) usage, but with some loss in accuracy \citep{moran2019energy}.

Many works have studied ReCA using the 5-bit memory benchmark. In \citep{nichele2017deep}, the structure of the CA was changed to a deep layered architecture and compared to a single layer, which resulted in noticeable performance improvements. In \citep{nichele2017reservoir}, the CA substrate was organised as consisting of two regions of different ECA rules. Different combinations of rules were explored, and some combinations showed great promise. In \citep{margem2019reservoir}, different methods of selection of cell history are used for the classification model and are tested on the 5-bit memory task, a temporal order task and arithmetic and logic operation tasks. 
In \citep{babson2019reservoir}, CA rules with multiple states and larger neighbourhoods were evolved and then tested on the 5-bit memory benchmark. In \citep{uragami2022universal}, ECA and asynchronous ECA are tested and compared on the 5-bit memory benchmark, mainly in the context of the distractor period.
In \citep{margem2020feed}, it was pointed out that the benchmark has no train test split. They modified the benchmark by training on just a few (2 or 3) of the 32 possible input streams, and some of the rules with more ordered behaviour could still solve this version of the benchmark. 
In \citep{glover2021dynamical}, the full ECA set was tested using key parameters of number of bits ($N_b$), redundancies ($R$) and Grid size. This work was also extended in \citep{glover2023investigating} to include more parameters such as Iterations ($I$) and Distractor Period ($D_p$). This paper also explained many of the unexpected results in the previous study, but perhaps as important, it similarly to \citep{margem2020feed} pointed out some weaknesses in the 5-bit memory benchmark. 

ReCA is explored on benchmarks other than the 5-bit memory benchmark. In \citep{moran2019energy, liang2021bloomca, garcia2020low}, ReCA is implemented (sometimes in hardware) and tested using MNIST. In \citep{mcdonald2017reservoir}, solved tasks of sine and square wave classification non-linear channel equalization, Santa Fe Laser Data and iris classification.

In \citep{kantic2024relicada}, a method for Rule Selection for ReCA was presented. Limiting the search space to only linear rules that obey a list of specific mathematical properties (see paper for details), the paper demonstrates that the method selects for rules in the high performance (95-80 percentile) bracket on several time-series prediction benchmarks compared to the entire Linear CA space of same neighbourhood and number of states.  

\section{Methodology}
\subsection{MNIST and bMNIST datasets}
Perhaps the most well-known benchmark in AI is MNIST \citep{bottou1994comparison}. It originally contained binary handwritten digits. However, it was grey-scaled to reduce the effect of aliasing. 

Here, as shown in Figure \ref{fig:bMNIST}, we simplify MNIST further by making the values of each image binary by simply rounding to 0 or 1 for each pixel, creating a binary MNIST (bMNIST). 
We adopt this approach not because we believe it's the most effective way to preserve information in terms of encoding. Rather, we choose it for its simplicity, as it allows for the most straightforward encoding strategy we could devise.

It's important to note that as we remove some of the information in the MNIST to make the bMNIST, any solution in terms of performance on the bMNIST is also applicable to the MNIST. However, it's crucial to understand that any solution developed for the MNIST may not necessarily be valid for the bMNIST. 

\begin{figure}[h]
\begin{center}
\includegraphics[width=0.6\columnwidth]{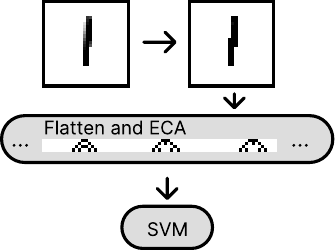} 
\caption{Example of MNIST ReCA solution}
\label{fig:ReCA MNIST}
\end{center}
\end{figure}

To encode the bMNIST into the CA we simply flatten it and set it as the initial condition of the CA. We acknowledge that flatting only retains relational information in the x plane and not in the y plane, but it is done for the sake of simplicity.

The MNIST dataset consists of 70000 labelled images of 28x28 pixels. We keep 33\%  in a holdout set, intended to be used later. The remaining 46900 are split into a 90\%/10\% train test split.


\subsection{UCR dataset}
The UCR Time Series Classification Archive \citep{UCRArchive2018} is a set of several smaller datasets that are time-series classification tasks. It is an improved version that considers criticisms of the previous UCR archive \citep{dau2019ucr}, such as a lack of benchmarks with varied length time-series. The authors point out some salient criticisms on the field of AI itself and give recommendations on best using the UCR datasets and other datasets, e.g., the importance of ablation testing when introducing multiple working parts.

\subsection{Similarity preserving expanded encoding (SimExp)}
\begin{figure}[h]
\begin{center}
\includegraphics[width=1\columnwidth]{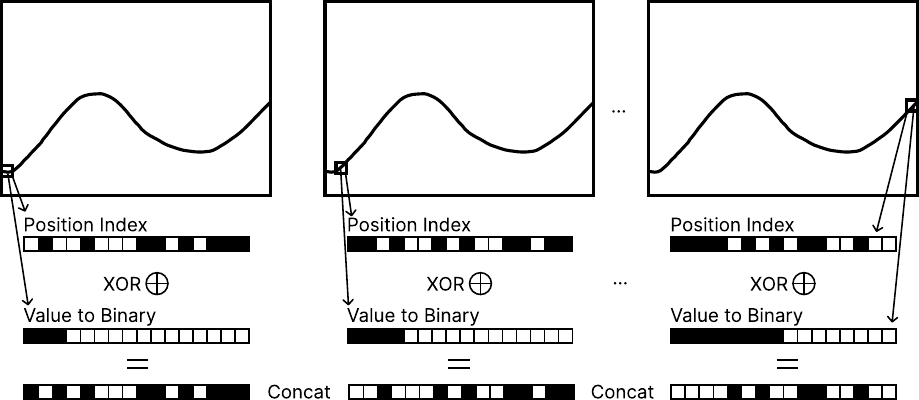} 
\caption{Example of time series being expanded into binary, without losing similarity. }
\label{fig:simExpExample}
\end{center}
\end{figure}

Binarising the MNIST dataset via rounding retains much of the useful information, but the same is not true for the UCR datasets. We used a similarity-preserved encoding that converts the data into a higher-dimensional binary vector, see Figure \ref{fig:simExpExample}. First, we make a random vector $n$ times larger than the largest vector in the dataset, representing a unique fixed value for every index position in the dataset's vectors. Then, we normalise the dataset between 0 and 1. Each value in a dataset vector was transformed into an $n$-sized vector, with 0s and 1s representing the float value (e.g., $n = 4,  0.75 = 1110$). Both binary vectors are then XORed together.  These values were concatenated into a vector, resulting in a vector n times larger than the original vector, which is binary. In our case, we used expansions of $n = \{16,32,64\}$, labelled as SimExp\{n\}, e.g., SimExp16. For this example, a vector of 10 float values would be transformed into 160 binary values.

Note that typically, this method does not concatenate every $n$-sized vector but rather embeds all the $n$-sized vectors into a binary cube of the same size as n for the sake of dimensionality reduction. This type of method has many names, but it is sometimes called Binary Embedding \citep{yi2015binary}. We call it something else to highlight the difference in terms of dimensionality reduction vs expansion. In our case, we started with concatenation as embedding into the cube might create some loss of information if the randomised index vectors were unlucky or the cube size was too small to contain the information. This way, one could have a baseline and, in a sense, another form of ablation test. As the method did not work well with our intended reservoir, the binary embedding method was not performed as intended. We still report it for the sake of scientific integrity but also because the method worked surprisingly well on its own.

\subsection{CA Reservoir and Classifier}
In all experiments reported, the encoded vector is set as the initial condition of an ECA of the same size as the vector. 3 CA steps are computed and concatenated to the initial condition. The 3 CA steps and the initial condition are then used to train a classifier, typically SVM with a linear kernel unless specified. Source code and results can be found at \citep{sourceCode}.

We tested briefly with other alternative number of CA steps to see if it would greatly impact performance. We kept the original parameters of 3 as the variation yielded no significant differences.

\section{Results}

\subsection{bMNIST}
\begin{figure}[h]
\centering
\includegraphics[width=1\textwidth]{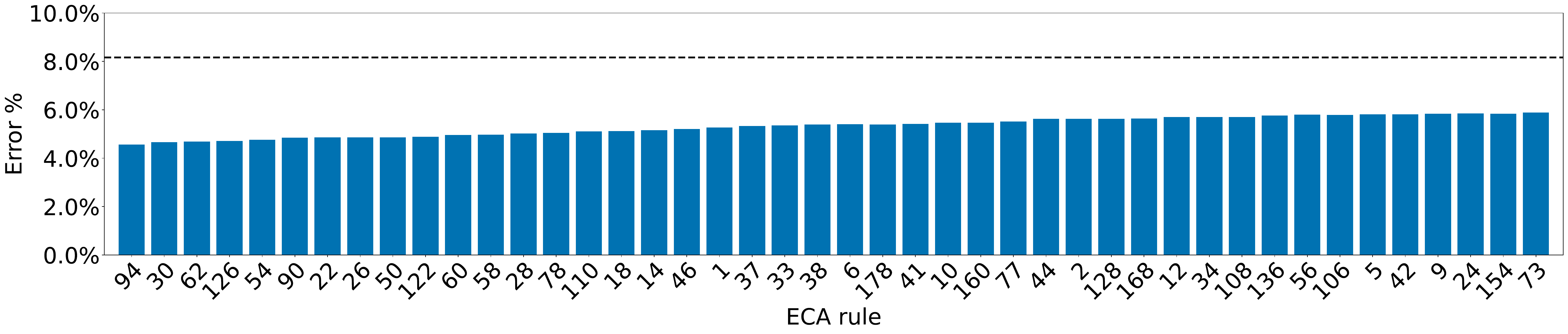} 
\includegraphics[width=1\textwidth]{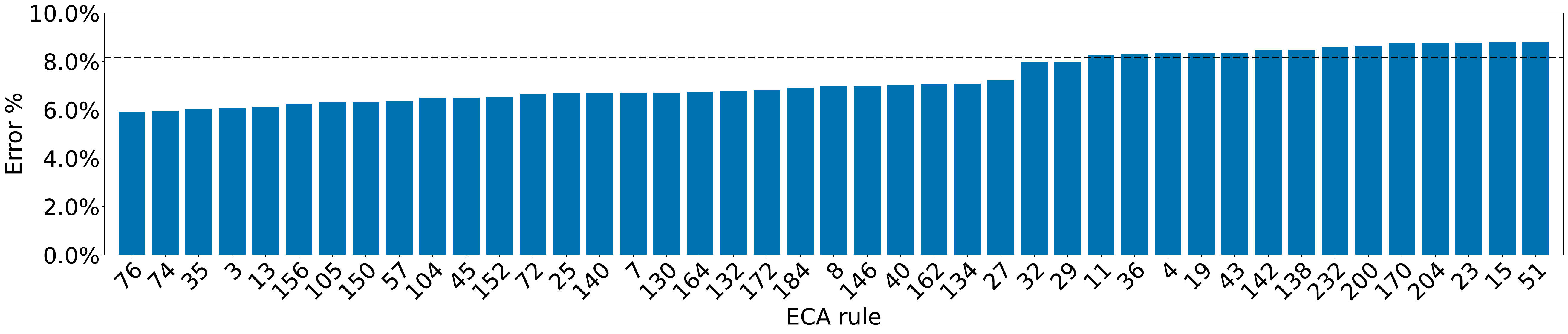} 
\caption{bMNIST error rate results, the dashed line represents an SVM solution without ECA. Rule 104 (in the second graph) is the first rule that is not statistically significant according to a Bonferroni-corrected t-test compared to the null hypothesis.}
\label{fig:bMNIST}
\end{figure}

As mentioned, ReCA has been tested on MNIST before. Both \citep{garcia2020low, moran2018reservoir} explored it with much more sophisticated methods and got better accuracy than our experiment (although they are not too far away). It is important to note that we will likely remove some useful information (though less than expected) from the benchmark by turning it into bMNIST. This makes it a different benchmark. Furthermore, we aim to make a minimal and simple solution, not to achieve the best performance.
As shown in Figure \ref{fig:bMNIST}, most ECA improve the performance significantly compared to no ECA Reservoir. The best rule (94) nearly halves the error rate from 8.2\% to 4.6\%. We also see that a wide range of rules improves the results. This improvement indicates that a parallel non-uniform design \citep{nichele2017reservoir} might be worth investigating for the bMNIST benchmark. 

We also briefly tested on the digits benchmark from SciKit-learn \citep{pedregosa2011scikit}, a smaller version of MNIST where the images are 8x8 pixels. This Digits benchmark was also binarised the same way as the bMNIST benchmark. ReCA worked well for this benchmark, but fewer rules performed well, and overall, the performance impact of the ECA was smaller. 

Additionally, we briefly tried other classifiers on the bMNIST benchmark; we used Rule 126 because it worked well on the binary Digits and the bMNIST benchmarks. The results can be seen in Table \ref{tab:bMNIST_other_classifiers}. The results show that CA also worked with an SVM using a Radial basis function (RBF) kernel. RBF kernel is not linear. Therefore, it is not typically used together with RC. The RBF kernel default parameters were used, and no hyperparameter tuning was performed. Therefore, we can not rule out that, for some parameters, the ECA could be redundant. We hypothesised that the ECA would only improve performance if the classifier had some form of selection of individual cells/pixels, allowing it to filter out the noise. Therefore, it was surprising that the method worked with a simple centroid classifier.
In contrast KNN ($n=5$) the ECA Reservoir made the performance significantly worse, which was more in line with our hypothesis. It worked with a single perceptron layer but did not improve performance when paired with larger multiple-layered perceptrons. This was expected as RC is a more energy-efficient replacement for a trained model.  
The multi-layered perceptron model received the highest performance of any classifier, which is unsurprising considering it had to be, in contrast, trained longer to settle the NN. We do not expect this to be the highest possible score, as we did close to no hyperparameter tuning. Therefore, it should not be hard to achieve a higher score. One can view this result as being closer to the theoretical limit of what can be done with the bMNIST benchmark, as we expect some information to have been lost in the binarisation, getting scores such as 99.83 \citep{byerly2021no}, might be impossible. Even so, this is surprisingly high, considering the binarisation reduced the precision in the data from 256 to 2.

\begin{table}[h]
    \centering
    \caption{Results of other classifying models for bMNIST using rule 126. "Improvement" is the improved performance contrasted with not using ECA. "Improvement \%" is the improved performance as a percentage of the remaining error rate. Essentially, 0 error rate would mean 100\% Improvement \% }
    \begin{tabular}{|c|c|c|c|c|}
    \hline
    Classifier & Accuracy & Error Rate & Improvement & Improvement \% \\
    \hline
         SVM linear & 95.29\% & 4.71\% & 3.45\% & 42.28\%\\
         SVM RBF & 95.63\% & 4.37\% & 2.37\% & 35.16\%\\
         Centroid & 75.62\% & 24.38\% & 5.70\% & 18.95\%\\
         KNN ($n=5$) & 90.59\% & 9.41\% & -1.65\% & -20.26\%\\
         single perceptron layer & 93.55\% & 6.45\% & 2.85\% & 30.65\%\\
         4 perceptron layers & 97.31\% & 2.69\% & -0.05\% & -1.82\%\\
         \hline
    \end{tabular}
    
    \label{tab:bMNIST_other_classifiers}
\end{table}

\subsection{UCR}

\begin{figure}[h]
\centering
\includegraphics[width=1\textwidth]{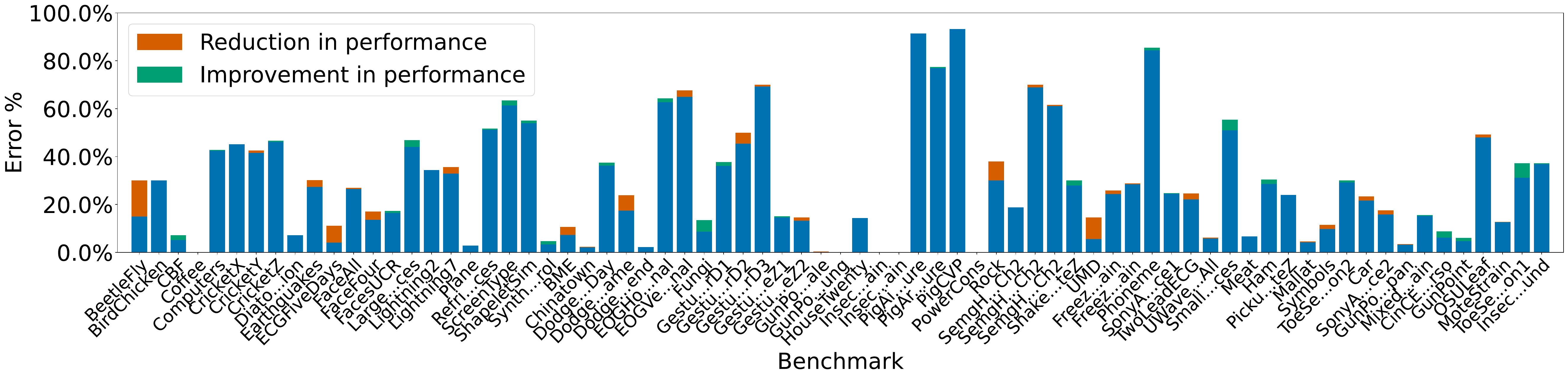} 
\includegraphics[width=1\textwidth]{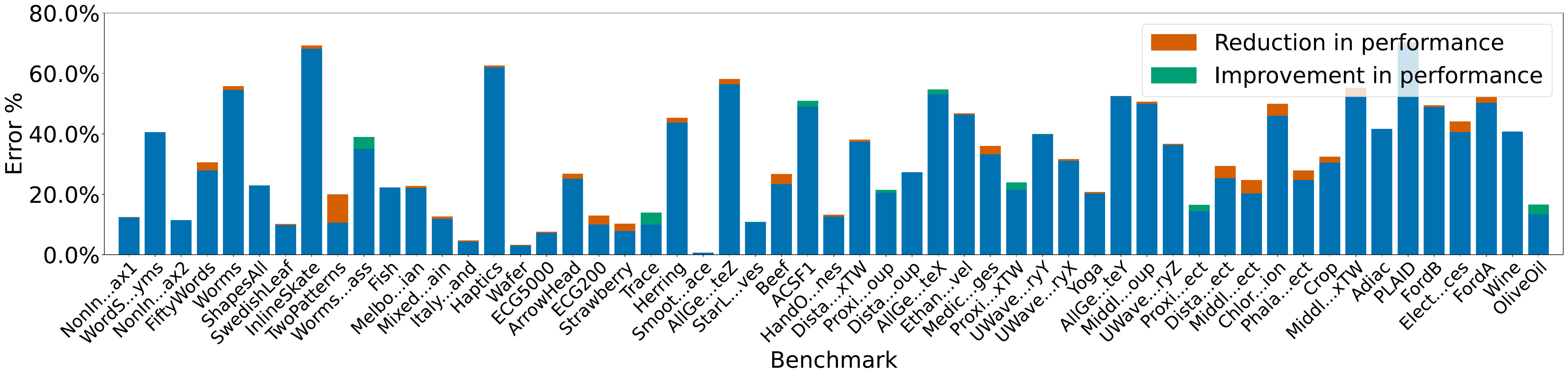} 

\caption{UCR dataset results with Rule 90 ECA Reservoir contrasted with its ablation. 
The first graph features all the benchmarks where the training set is perfectly solvable with an SVM with a linear kernel (linearly solvable). The second graph features the benchmarks that were not linearly solvable and are sorted from best to worst based on how well an SVM with a linear kernel solved them.  
}
\label{fig:ucr_SimExpECASVM}    
\end{figure}

\begin{figure}[h]
    \centering
    \includegraphics[width=1\textwidth]{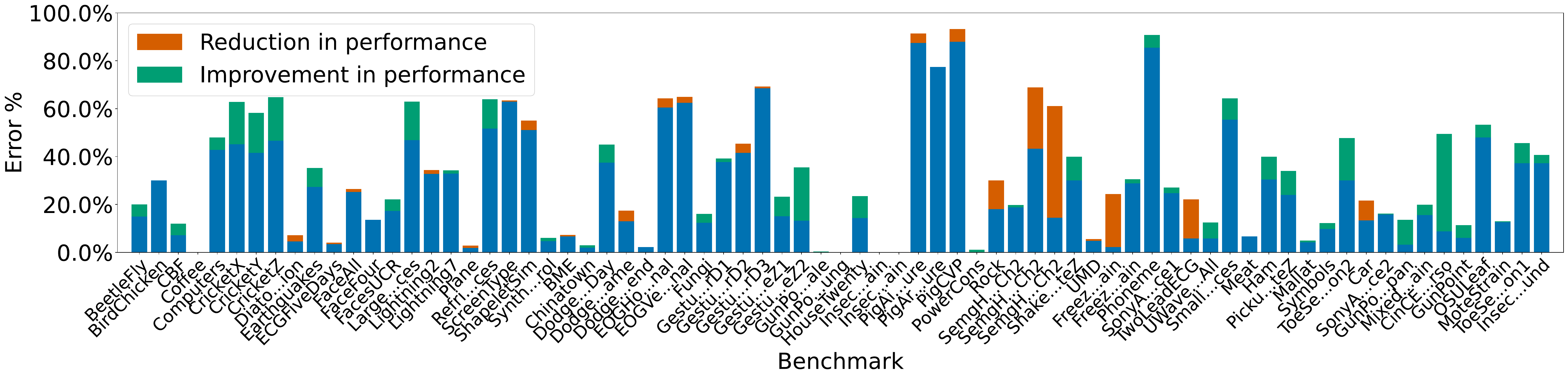} 
    \includegraphics[width=1\textwidth]{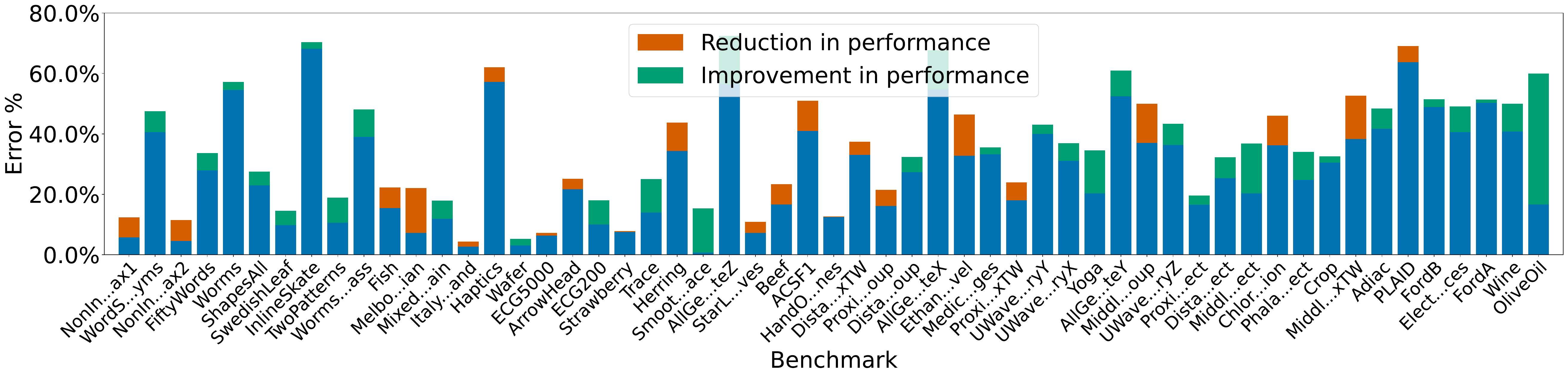} 
        \caption{Same as Figure \ref{fig:ucr_SimExpECASVM}, but the SimExp16 experiment contrasted with its ablation.}
\label{fig:ucr_SimExpSVM}
\end{figure}

In Figure \ref{fig:ucr_SimExpECASVM}, Rule 90 is shown. While it is performing relatively well considering the simplicity of the solution, when compared directly to the same solution without the ECA steps, it is, in fact, performing better on average. This result means that the ECA part of the ReCA solution is, in general, functionally useless in this implementation, though not necessarily for individual datasets. 
None of the ECA rules manged to overall improve results, 90 was picked as an example due to its popularity in ReCA. 
In contrast, the SimExp16 method works better than just a Linear SVM, as seen in Figure \ref{fig:ucr_SimExpSVM}. This difference becomes more significant in SimExp32 and furthermore in SimExp64, where it got 4.6\% lower error on average than SVM with a linear kernel; this is shown in Table \ref{tab:ucr_SimExp}. Additionally, as we trained the SVM ablation test on the original benchmark where the numbers are floating point values, we essentially would have some loss of precision even in the SimExp64 as it only has 64 precision values in the encoding, meaning it is on par for precision with a 6-bit integer for precision. 
We split the benchmarks into linear solvable and not linearly solvable because we hypothesised that in benchmarks where the training data was already linearly separable, the SVM would find no benefit from the ECA-developed features. We, therefore, predicted no performance gain from the first graph in Figure \ref{fig:ucr_SimExpECASVM}. This prediction was technically confirmed, but as we saw no benefit from the second graph, the prediction should be viewed in spirit as false for this experiment. In \citep{dau2019ucr}, it was recommended to use a Texas sharpshooter plot, as a model is only good if one can predict what benchmarks it will perform well on or if it improves overall performance. We instead used a more straightforward bar graph, and as the hypothesis failed, there was no need to prove the hypothesis in the graph. A bar graph allows for a better overview. 

Interestingly, we also observed that the SimExp64 performed marginally better than dynamic time warping (DTW) (W=100), with a mean error rate reduction of $0.54\% $; this might be low, but it warrants further study.

\begin{table}[h]
    \centering
    \caption{Average UCR archive results using only SimExp and SVM}
    \begin{tabular}{|c|c|c|c|}
    \hline
    Expansion size & Accuracy & error rate & Improvement\\
    \hline
         SimExp16 & 70.79\% & 29.21\% & 2.15\%\\
         SimExp32 & 72.37\% & 27.63\% & 3.73\%\\
         SimExp64 & 73.32\% & 26.68\% & 4.68\%\\
         \hline
    \end{tabular}
    
    \label{tab:ucr_SimExp}
\end{table}

\section{Discussion}
\subsection{Locality vs Globality}
Considering that in the UCR dataset, the DTW method does well, and the DTW essentially allows global properties of a time series to be non-linearly mapped and compared \citep{dau2019ucr}, we would assume that global properties of the datasets are essential for performance. If we contrast this to our 3 CA step solution for bMNIST, global features are less crucial for bMNIST, as CA's speed of light means that no information can move beyond the 3 nearest cells in the CA. Considering these two observations, we argue that ReCA works best for local features and poorly for global features. As most ReCA explore temporal problems or problems that can be solved with local features, an open question is how to configure ReCA to handle global features better. 
\subsection{MNIST vs bMNIST}
Our results indicate that turning MNIST into bMNIST can be done trivially, reducing the precision of the data from 256 (uint8) to 2 (binary) while losing accuracy capability of at most 2.69\% (probably much less) as MNIST can be reduced to $\frac{1}{8}$ of it memory size and $\frac{1}{128}$ of its precision, without losing much accuracy. This can have implications for any work that binarises the MNIST dataset. Essentially, there is not much necessary or non-redundant information encoded into the grayscale dimension, so any work that tries to preserve it might naturally assume the methods used are more successful than they are. 

Quantization is a common practice to reduce the precision of values in datasets, e.g., float64 to float16. Perhaps many other benchmarks can also be reduced to binary. This would be great news for binary models such as CA. 

\subsection{Deception of good encoding}
SimExp works very well for preserving useful information in the UCR datasets, while combining it with ECA did not improve the results. This showcases the value of ablation tests. Otherwise, we might have mistakenly claimed that ReCA is effective for UCR classification when, in fact, it is the SimExp encoding that does all the work. Why it is so effective is a question that remains open for further research.



\section{Conclusion and Future Work}
We have demonstrated a simple ReCA system in action. We have also shown a deceptive example where ReCA might seem to work but the actual work is done by the encoding scheme, highlighting the importance of ablation testing. 

A possibility for future work would be a comparison with a ReCA system using linear regression or ridge regression readouts instead of SVM. A more difficult task would allow rigorous identification of suitable methods for encoding global features into the CA. We also have left 33\% of MNIST in a holdout set for further use. Considering the relatively good results, a more sophisticated encoding method, e.g., including the y-dimension of MNIST would be prudent. Additionally, the promising results of the SimExp encoding warrant further study.

\begin{credits}
\subsubsection{\ackname} This work was partially financed by the Research Council of Norway’s DeepCA project, grant agreement 286558.

\subsubsection{\discintname}
The authors have no competing interests.
\end{credits}

\bibliographystyle{splncs04}
\bibliography{example} 

%
%
%
%




\end{document}